%% file: iclr2027_conference.tex
\documentclass{article}

\PassOptionsToPackage{dvipsnames}{xcolor}
\usepackage{iclr2027_conference,times}

\usepackage[utf8]{inputenc}
\usepackage[T1]{fontenc}
\usepackage{hyperref}
\usepackage{url}
\usepackage{booktabs}
\usepackage{amsfonts}
\usepackage{nicefrac}
\usepackage{microtype}
\usepackage{float}
\usepackage{adjustbox}
\usepackage{tabularx}
\usepackage{multirow}
\usepackage{mdframed}
\usepackage{graphicx}
\usepackage{wrapfig}
\usepackage{subcaption}
\usepackage{amsmath}
\usepackage[capitalize,noabbrev]{cleveref}
\usepackage{enumitem}
\usepackage{bbm}

\title{LLM Complexity: Towards a Computational Complexity View of Large Language Model Inference}

\author{%
  Yu-Hung Wu$^{1}$\thanks{Equal contribution} \quad
  Shao-Xiang Yuan$^{2}$\footnotemark[1] \quad
  Chan-Hung Yu$^{1}$
  \quad
  Zih-Ching Chen$^{3}$ \\
  \textbf{Kuan-Hao Huang$^{4}$ \quad
  Shao-Hua Sun$^{1, 2, 5}$} \\
  $^{1}$Graduate Institute of Communication Engineering, National Taiwan University (NTU)\\
  $^{2}$Department of Electrical Engineering, NTU\\
  $^{3}$NVIDIA AI Technology Center, NVIDIA Corporation, Santa Clara, CA, USA\\
  $^{4}$Texas A\&M University\\
  $^{5}$NTU Artificial Intelligence Center of Research Excellence
}

\iclrfinalcopy % Uncomment for camera-ready version, but NOT for submission.

\begin{document}

\input{macro}

\maketitle

\begin{abstract}
  \input{sections/abstract}
\end{abstract}

\input{sections/intro}
\input{sections/related_work}
\input{sections/method}
\input{sections/LLM_Complexity}
\input{sections/complexity_results}

\input{sections/curriculum}
\input{sections/traditional_complexity}
\input{sections/real_world}
\input{sections/conclusion}

% ICLR 2027 requires an AI use statement before the references. Add the
% authors' reviewed disclosure here before submission; do not copy the generic
% template wording without adapting it to the actual use of generative AI.

\bibliography{citations}
\bibliographystyle{iclr2027_conference}

\appendix
\input{sections/appendix/main}

\end{document}

%% file: macro.tex
\newcommand{\mytitle}{}
\newcommand{\myshorttitle}{\mytitle}
\newcommand{\method}{LLM Complexity}

% abbr
\newcommand{\etal}{\textit{et al}.}
\newcommand{\ie}{\textit{i}.\textit{e}.,\ }
\newcommand{\eg}{\textit{e}.\textit{g}.,\ }

% ref
\newcommand{\myfig}[1]{Figure \ref{#1}}
\newcommand{\mytable}[1]{Table \ref{#1}}
\newcommand{\myeq}[1]{Equation \ref{#1}}
\newcommand{\mysec}[1]{Section \ref{#1}}
\newcommand{\myalgo}[1]{Algorithm \ref{#1}}
\newcommand{\myapp}[1]{Appendix \ref{#1}}

% % section & paragraph
\newcommand{\myparagraph}[1]{\noindent \textbf{#1.}}
\newcommand{\vspacesection}[1]{\vspace{-0.12cm}
\section{#1}
\vspace{-0.12cm}}

\newcommand{\vspacesubsection}[1]{\vspace{-0.12cm}
\subsection{#1}
\vspace{-0.12cm}}

\newcommand{\vspacesubsubsection}[1]{\vspace{-0.12cm}
\subsubsection{\textbf{#1}}
\vspace{-0.12cm}}

%% file: sections/abstract.tex
Computational complexity theory offers a principled framework for characterizing the efficiency of computational problems, and Large Language Models (LLMs) can similarly be viewed as computational systems executing a structured process of input, inference-time compute, and output. In this context, the number of generated tokens emerges as a natural and directly observable proxy for inference-time compute. However, we show that the tempting analogy between LLM token length and classical time complexity does not hold. We disprove the hypothesis that inference-time token usage aligns monotonically with classical time complexity, demonstrating a fundamental divergence grounded in how LLMs actually allocate computation during inference. To this end, we introduce \method{}, a new metric designed to quantify problem complexity from the perspective of LLMs by characterizing the area bounded above by the accuracy–token curve. We find that diverse LLMs induce highly consistent task rankings under \method{}, indicating that the metric captures stable notions of difficulty across models. We leverage this property to construct an easy-to-hard curriculum for supervised fine-tuning, which improves accuracy while reducing token consumption.
Finally, we demonstrate that \method{} generalizes to real-world tasks, producing more stable cross-model correlations than human-annotated difficulty labels.

%% file: sections/intro.tex
\vspacesection{Introduction}

Computational complexity theory lies at the core of computer science, offering a principled framework to characterize the efficiency and intrinsic properties of computational problems~\citep{10.1145/800157.805047, Karp1972, arora2006computational}.
Traditionally, complexity is defined in terms of the asymptotic growth of the required computational resources, \eg time complexity, as a function of problem size, measured by the number of elementary computational steps~\citep{hartmanis1965computational, sipser1996introductioncomputation}.
Large language models (LLMs), despite their architectural and operational differences from classical algorithms, can also be viewed as computational systems executing a structured process of the form: ``input $\rightarrow$ inference-time compute $\rightarrow$ output.''
This perspective naturally raises a fundamental question: to what extent \emph{can the behavior and capabilities of LLMs be analyzed through the lens of computational complexity theory?}

Addressing this question requires defining an appropriate notion of ``elementary operations'', analogous to those used in time complexity, for inference-time compute in LLMs.
During generation, an LLM produces tokens autoregressively, with each token representing an intermediate computational state that later steps condition on~\citep {Vaswani2017attention, brown2020fewshotgpt3}.
As a result, the number of generated tokens emerges as a natural and directly observable proxy for the inference-time compute.
Moreover, a growing body of evidence suggests that allocating more inference-time compute—often manifested as longer reasoning traces—enables LLMs to solve a broader class of tasks~\citep{jaech2024openai, comanici2025gemini, zhang2025system1}.
These observations invite a tempting analogy: \emph{does LLM token length play a role analogous to time complexity in classical computation?}

\input{images/intro_fig}

In this work, we show that this analogy may not hold. Specifically, we challenge the hypothesis that inference-time token usage monotonically aligns with classical worst-case time complexity through two counterexamples.
First, we compare LLM accuracy and token usage across classical algorithmic problems. As shown in \myfig{fig:introa}, \emph{sorting}, an $O(n \log n)$ problem, requires more tokens to achieve 50\% accuracy than the NP-hard \emph{subset sum}. This likely arises because \emph{sorting} requires long, structured outputs that amplify error propagation, whereas \emph{subset sum} can often be solved through localized reasoning.
Second, increasing problem size does not necessarily lead to monotonic performance degradation (\myfig{fig:introb}). In search-based problems, LLMs can often exploit heuristic reasoning to focus on a small subset of plausible candidates, making performance only weakly dependent on input size.
Together, these findings suggest that classical worst-case time complexity alone does not adequately characterize inference-time token usage in LLMs. This motivates an operational notion of complexity grounded in autoregressive inference, where generated tokens serve as the primitive computational step.

To this end, we introduce \method{}, a new metric designed to quantify problem complexity from the perspective of autoregressive large language models.
Rather than relying on classical computational complexity, we characterize each problem by its accuracy–token curve, which captures performance as a function of a token budget.
Intuitively, problems that achieve high accuracy under tight token constraints should be considered less complex, while those requiring substantially more tokens to reach comparable performance are more complex.
Formally, we define \method{} as the area bounded above by the accuracy–token curve: a smaller area corresponds to lower complexity, whereas a larger area indicates higher complexity due to greater requirements of inference-time compute.
This formulation yields a problem complexity measure that is both model-aware and operationally meaningful, and reveals a striking property: despite differences in architecture and scale, various LLMs tend to induce highly consistent rankings under \method{}.

We empirically evaluated this cross-model correlation by ranking twelve algorithmic tasks across five LLMs from three distinct model families.
Using Spearman's $\rho$ and Kendall's $\tau_{b}$, we find that \method{} rankings exhibit stronger agreement across models than rankings derived from common behavioral heuristics such as peak accuracy.
Also, we find that \method{} is robust to prompt variations, indicating that it captures stable task-level inference complexity rather than superficial signals \eg output token length.
%\method{} is robust to prompt variations, indicating that it captures intrinsic task complexity rather than superficial signals \eg output token length.
Exploiting this consistency, we demonstrate a practical application: complexity estimates obtained from a larger model can be used to construct an easy-to-hard curriculum for supervised fine-tuning a smaller model, yielding higher accuracy while reducing token consumption compared to random data ordering.
We also extend \method{} to real-world tasks on DeepMath~\cite{he2025deepmath} and RuleArena~\citep{zhou2024rulearena}, and show that it yields more stable cross-model correlations than human-annotated difficulty labels, suggesting that \method{} more faithfully captures how LLMs allocate inference-time compute during problem solving.

%% file: images/intro_fig.tex
\begin{wrapfigure}{r}{0.49\linewidth}
  \vspace{-8pt}
  \centering
  \begin{subfigure}[t]{0.485\linewidth}
    \centering
    \includegraphics[width=\linewidth]{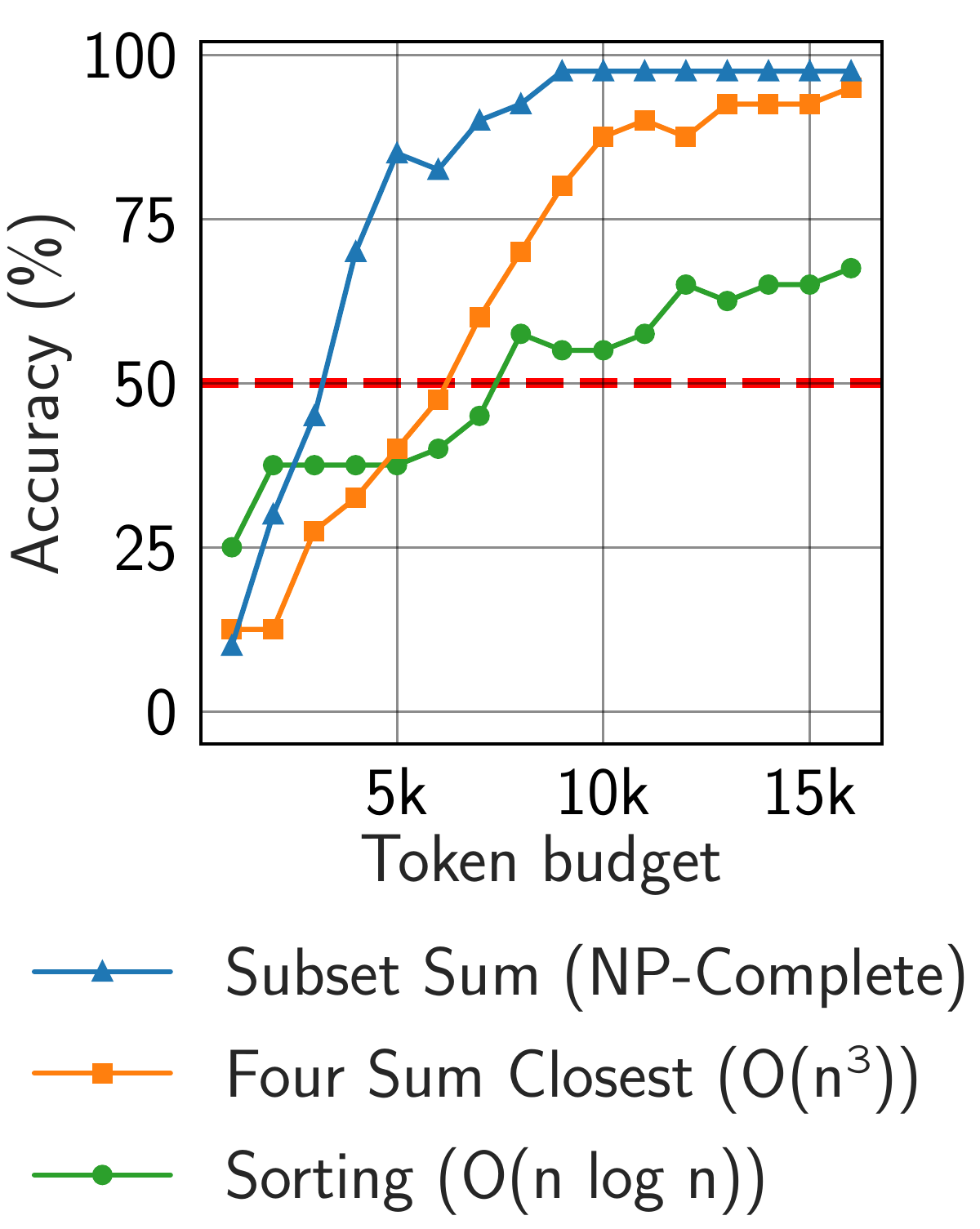}
    \caption{Token usage misaligns with worst-case time complexity.}
    \label{fig:introa}
  \end{subfigure}\hfill
  \begin{subfigure}[t]{0.485\linewidth}
    \centering
    \includegraphics[width=\linewidth]{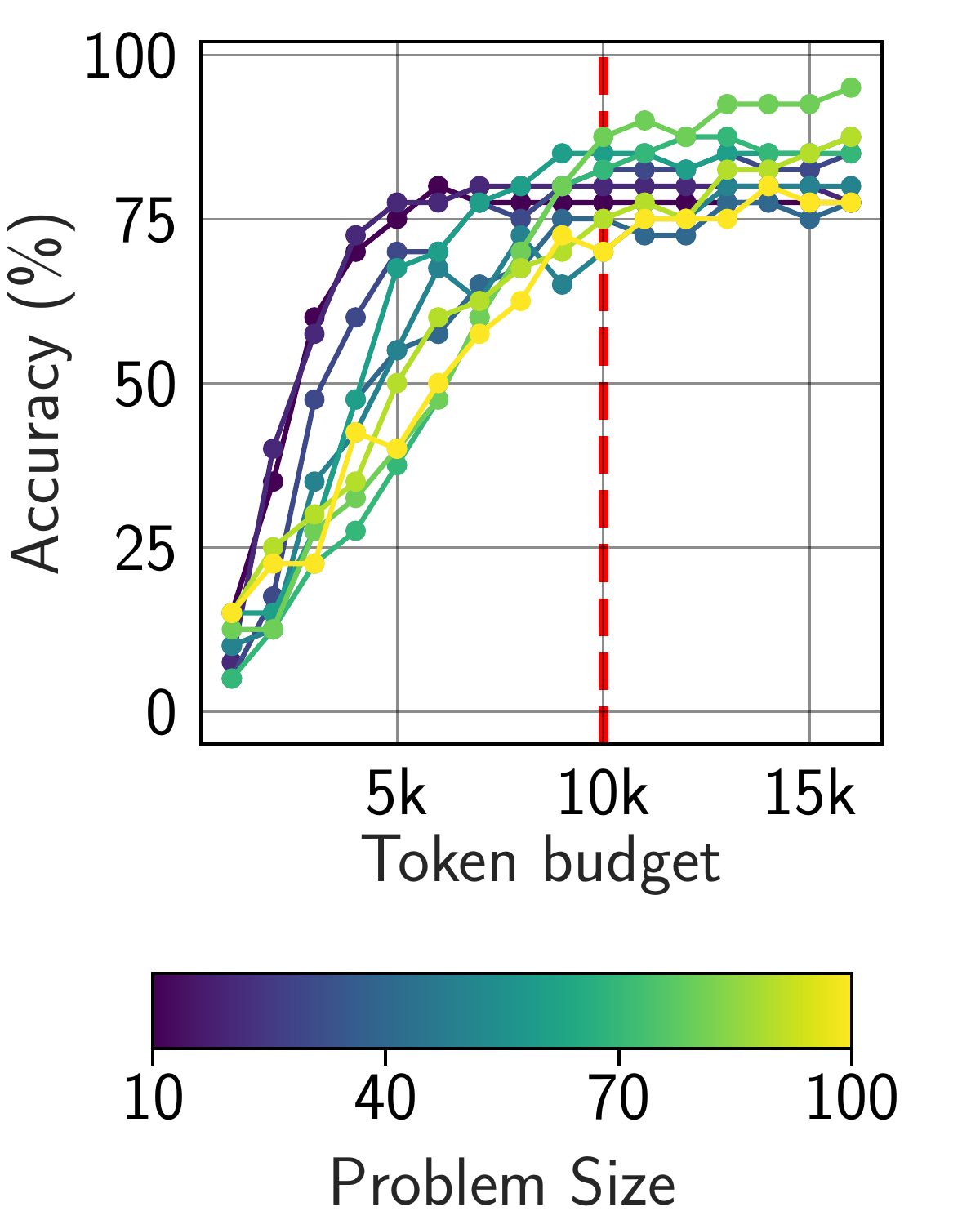}
    \caption{Inference difficulty is not monotonic with problem size.}
    \label{fig:introb}
  \end{subfigure}
  \caption{\textbf{Time complexity does not align with LLMs' inference-time token usage.} 
    We conduct experiments on \textsc{Nemotron-253B} and report the relationship between the token budget and accuracy. (a) For problem size 80, \emph{sorting} ($O(n \log n)$) requires more tokens to reach 50\% accuracy than some tasks with higher theoretical complexity. (b) For \emph{four sum closest}, accuracy at a 10K-token budget does not decrease monotonically with problem size. Together, these results highlight a mismatch between algorithmic complexity and LLM inference behavior.
  }
  \label{fig:intro}
  \vspace{-6pt}
\end{wrapfigure}

%% file: sections/related_work.tex
\vspacesection{Related work}

\noindent \textbf{LLMs as computational systems.}
\citet{strobl-etal-2024-formal} models Transformers and LLMs as computational systems and analyzes their expressivity using tools from computation and formal language theory. \citet{perez2021,bhattamishra-etal-2020-computational,jiang2025softmaxtransformersturingcomplete} further establishes computational universality.
Unlike these studies, we focus on an operational notion of inference-time cost of LLMs 
and analyze it through the lens of computational complexity theory.

\noindent \textbf{LLM reasoning and inference-time compute.}
LLM reasoning methods show that allocating more inference-time compute can substantially improve LLM performance. 
Chain-of-thought~\citep{Wei2022} prompting elicits explicit intermediate steps, and self-consistency scales inference-time compute by sampling multiple reasoning paths and aggregating to the most consistent answer~\citep{wang2022self}. 
Recent work further explores search/refinement-based scaling of inference-time compute, \eg Mind Evolution \citep{lee2025evolving}, alongside studies of redundant ``overthinking'' and budget-aware allocation \citep{chiang2024overreasoningredundantcalculationlarge, chen2025think23overthinkingo1like, damani2024learninghardthinkinputadaptive}. 
In contrast, we quantify how accuracy scales under a bounded token budget and distill this relationship into a task-level complexity score for cross-task comparison and curriculum design.

\noindent \textbf{On problem difficulty for LLMs.}
Prior work defines difficulty either \emph{externally}, from observed success across a population of solvers \citep{ding2024easy2hard,zhou2025lost}, or \emph{internally}, as model-perceived hardness inferred from a target LLM’s hidden representations \citep{zhu-etal-2025-llm}. 
This treats the LLM as a computational system and characterizes difficulty through a complexity-based lens, using inference-time token consumption as an operational resource measure to derive a cross-model-consistent notion of task difficulty.

%% file: sections/method.tex
\vspacesection{Token usage vs. time complexity}
\input{table/outlen_vs_timecorr}

When LLMs are viewed as computational systems, tokens provide a directly observable proxy for inference-time compute. This raises the hypothesis that token usage should increase monotonically with the computational time complexity of a problem. To test this, we evaluate five LLMs on twelve algorithmic tasks spanning diverse complexity classes 
and show that token usage does not correlate with worst-case time complexity across tasks.

\vspacesubsection{Experimental setup}
\label{sec:experimental_setup}

\noindent \textbf{Data construction.}
To enable controlled and comparable scaling analysis, we consider a suite of classical algorithmic problems, ranging from $O(n \log n)$ \emph{sorting} to NP-hard \emph{subset sum}. Each problem takes a list of length $n$ as input, where $n$ is the problem size. For evaluation, we generate 32 instances per size using task-specific generators with standardized formats, varying $n \in {10, 20, \ldots, 100}$. The full task list and prompts are provided in \Cref{appendix:task_table}.

\noindent \textbf{Models.}
We evaluate five open-weight LLMs: \textsc{Llama-3.3-nemotron-super-49B-v1} and \textsc{Llama-3.1-nemotron-ultra-253B-v1} from the Nemotron family~\cite{bercovich2025llamanemotronefficientreasoningmodels}, \textsc{gpt-oss-20b} and \textsc{gpt-oss-120b} from the gpt-oss family~\cite{openai2025gptoss120bgptoss20bmodel}, and \textsc{Qwen-235B} from the Qwen3 family~\cite{yang2025qwen3}.
We focus on open-weight models as our evaluation requires access to the full generated reasoning trace, where most closed-source APIs, \eg GPT, Claude, Gemini, expose only partial or summarized reasoning traces.

Note that all the experiments are conducted in a text-only setting. Models are not permitted to use external tools, programs, or execution environments; this is enforced via prompt design and, when available, model-specific configurations.

\vspacesubsection{Correlation between token usage \& time complexity}
\label{sec:correlation_with_time}

To test whether token usage reflects worst-case time complexity, we prompt each LLM to solve each task instance step by step and record the generated tokens. For each problem, we average output length across sizes, rank tasks by this statistic, and compare it to rankings induced by worst-case time complexity using Spearman’s $\rho$ and Kendall’s $\tau_b$.

The Results in~\mytable{tab:outlen_vs_timecorr} show uniformly low correlations across all five models, indicating that token usage is weakly aligned with worst-case time complexity. To further analyze this, we report average output length per task in \Cref{app:avg_outlen}. We make two observations. First, \emph{four sum}, \emph{four sum in range}, and \emph{three sum in range} show low token usage, as these search-style tasks enable heuristic reasoning without exhaustive exploration. Second, \emph{sorting}, despite its $O(n \log n)$ complexity, incurs high token usage due to long structured outputs that increase generation length and error propagation.

Overall, the results suggest that token usage reflects LLM-specific inference behaviors rather than algorithmic worst-case time complexity, highlighting the need for a complexity notion grounded in how LLMs allocate computation during inference.

%% file: table/outlen_vs_timecorr.tex
\begin{wraptable}[20]{r}{0.45 \linewidth}
  \vspace{-4pt}
  \centering
  \caption{
  \textbf{Weak alignment between worst-case algorithmic time complexity and LLM token usage.} 
  We conduct the experiment with 12 algorithmic problems, each evaluated across 10 problem sizes.
  For each model, we compute the correlation between the average output length and a task-level time-complexity ranking derived from tight, widely accepted worst-case bounds (\eg \emph{sorting} with $O(n \log n)$).
  Across all five models, both Spearman’s $\rho$ and Kendall’s $\tau_{b}$ remain small, indicating weak alignment between inference-time token usage and worst-case algorithmic time complexity.
  }
  \label{tab:outlen_vs_timecorr}
  \resizebox{\linewidth}{!}{
  \begin{tabular}{lc}
    \toprule
    \textbf{Model} & \textbf{Spearman's $\rho$ / Kendall's $\tau_{b}$} \\
    \midrule
    \textsc{Nemotron-49B}   & 0.102 / 0.103 \\
    \textsc{Nemotron-253B}  & 0.008 / 0.098 \\
    \textsc{gpt-oss-20b}    & 0.213 / 0.165 \\
    \textsc{gpt-oss-120b}   & 0.213 / 0.165 \\
    \textsc{Qwen-235B}      & 0.188 / 0.132 \\
    \bottomrule
  \end{tabular}}
\end{wraptable}

%% file: sections/LLM_Complexity.tex
\vspacesection{Approach}

Since  time complexity does not directly characterize LLM reasoning behavior, we seek an effective metric to quantify task complexity under generation constraints with the following desiderata:

\begin{enumerate}[leftmargin=15pt,itemsep=1pt,topsep=0pt]

\item \textbf{Cross-model agreement.} The metric should induce consistent complexity rankings across various models and across algorithmic and real-world tasks, rather than model-dependent behaviors.

\item \textbf{Prompt robustness.} The metric should remain stable under different prompt variations, rather than being driven by prompt-dependent behaviors, \eg verbosity or reasoning style.

\item \textbf{Learning utility.} The metric should provide practical value for downstream applications, such as constructing more effective and efficient curriculum learning strategies.

\end{enumerate}

To this end, we describe our protocol for measuring performance across varying token budgets (\mysec{method:inference}), and then formalize \method{}, our proposed metric, as a scalar measure of the computational cost required to achieve high accuracy (\mysec{method:complexity}).

\vspacesubsection{Calculating accuracy across token budgets}
\label{method:inference}

To consistently measure performance across budgets, we adopt a prefix-truncation procedure~\cite{lee2025llmscompresschainofthoughttoken, ghosal2025doesthinkinghelpmirage, muennighoff-etal-2025-s1}, ensuring that differences reflect marginal compute rather than sampling variance. We define token budgets $\mathcal{B}={1{,}000,\dots,16{,}000}$ and generate a single long reasoning trace per instance with $b_{\max}=16{,}000$, using continuation prompts if needed (\Cref{appendix:continue-prompt}). Following “budget forcing”~\citep{muennighoff-etal-2025-s1}, we truncate the trace at each $b \in \mathcal{B}$ and prompt the model to answer from the partial trace.

We evaluate different truncations of the same reasoning process across budgets, using task accuracy at each budget $b$. Per-instance correctness is defined as $\mathbf{1}[\hat{y}=y]$, where $y$ is the ground truth and $\hat{y}=\hat{y}(x;c,n,b)$ is the final answer under budget $b$. For each problem $c$ and size $n$, instances $x \sim \mathcal{D}_{c,n}$, and accuracy at budget $b$ is the expected correctness:
$\mathrm{Acc}(c,n,b)
=
\mathbb{E}_{x \sim \mathcal{D}_{c,n}}
\left[
\mathbf{1}\big[\hat{y}(x;c,n,b)=y(x;c,n)\big]
\right].$
Evaluating $\mathrm{Acc}(c,n,b)$ over the budget set $\{b_1,\ldots,b_K\}$ yields the performance measurements used in the following section.

\vspacesubsection{Defining \method{}}
\label{method:complexity}
\input{images/inference_utility}

A robust characterization of task complexity should reflect how accuracy changes as the available token budget varies. Therefore, we evaluate each task under multiple token budgets (\mysec{method:inference}) and define \method{} by aggregating the resulting budget–accuracy measurements.

As illustrated in \cref{fig:inference_utility}, we summarize each task’s budget--accuracy scaling behavior using the \textbf{inference utility}, defined as the area under the accuracy--budget curve over $\mathcal{B}$. We compute this area using the trapezoidal rule in log-budget space:
$
\mathcal{U}(c,n) =\sum_{i=1}^{N-1}
\frac{\text{Acc}(c,n,\log{b_i}) + \text{Acc}(c,n,\log{b_{i+1}})}{2} \cdot \left(\log{b_{i+1}} - \log{b_i}\right).$
Using log-budget space prevents a large token budget from disproportionately dominating the aggregate.

To facilitate comparison across tasks and problem sizes, we normalize by the log-budget interval length:
$\bar{\mathcal{U}}(c,n)
\;=\;
\frac{\mathcal{U}(c,n)}
{\log b_{\max} - \log b_{\min}},\qquad
\bar{\mathcal{U}}(c,n) \in [0,1],
$
where $b_{\min}=b_1$ and $b_{\max}=b_N$.

Finally, we define the \method{} of task $c$ at problem size $n$ as
$\mathcal{C}(c,n) = 1 - \bar{\mathcal{U}}(c,n)$, so larger values of $\mathcal{C}(c,n)$ indicate that achieving high accuracy requires a greater inference-time token budget.
\method{} provides a cross-task comparable notion of complexity under bounded generation. It is based on inference utility, which aggregates accuracy across multiple token budgets in log-budget space. We examine its key properties in the following sections. Moreover, \method{} is insensitive to tokenizer choice: small variations typically induce near-uniform rescaling of token counts, which does not materially affect the resulting rankings.

%% file: images/inference_utility.tex
% \begin{figure}[t]
%   \centering
%   \begin{minipage}{0.58\columnwidth}
%     \centering
%     \includegraphics[width=\linewidth]{images/inference_utility_example}
%   \end{minipage}
%   \hfill
%   \begin{minipage}{0.40\columnwidth}
%     \vspace{-17pt}
%     \caption{
%       \textbf{Illustration of inference utility}. 
%       We plot accuracy as a function of the token budget. The shaded area under the curve represents aggregate accuracy across the budget.
%     }
%     \label{fig:inference_utility}
%   \end{minipage}
% \end{figure}

\begin{wrapfigure}[8]{H}{0.5\linewidth}
\vspace{-20pt}
  \centering
  \begin{minipage}{0.52\linewidth}
    \centering
    \includegraphics[width=\linewidth]{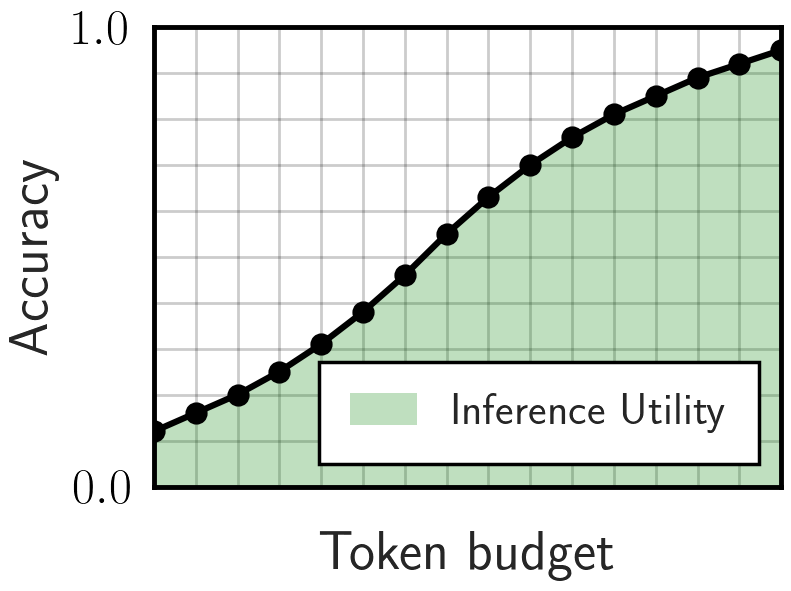}
  \end{minipage}\hfill
  \begin{minipage}{0.45\linewidth}
    \caption{\textbf{Illustration of inference utility.}
    We plot accuracy as a function of token budget. The shaded area under the curve represents aggregate accuracy across the budget.}
    \label{fig:inference_utility}
  \end{minipage}
\end{wrapfigure}

%% file: sections/complexity_results.tex
\input{table/correlation_cross_model}

\vspacesection{Experimental results}
\label{sec:main_results}

\input{images/avg_complexity_score}

To determine whether \method{} captures an effective complexity signal, we first map the \emph{topology of complexity} across tasks and problem sizes (\Cref{subsec:Complexity_Across_Tasks_and_Problem_Sizes}).
Then, we evaluate its \textbf{cross-model agreement} through correlations across different models and tasks (\Cref{subsec:Cross-Model_Correlation_of_Complexity}). 
Finally, we assess its \textbf{prompt robustness}, examining whether the metric remains stable under prompt variations rather than being driven by superficial prompt-dependent behaviors such as verbosity or reasoning style (\Cref{subsec:Prompt_Sensitivity_of_Complexity}).

Note that we focus on open-weight models as our evaluation requires access to the full generated reasoning trace, where most closed-source APIs, \eg GPT, Claude, Gemini, expose only partial or summarized reasoning traces.

\vspacesubsection{\method{} across tasks}
\label{subsec:Complexity_Across_Tasks_and_Problem_Sizes}

To characterize the topology of task complexity, Figure~\ref{fig:avg_complexity_score} shows the average \method{} across tasks and problem sizes.
Tasks with lower complexity exhibit stable scores across input lengths, whereas higher-complexity tasks show increasing complexity as problem size grows. This suggests that harder tasks require tracking and updating more intermediate states, leading to higher inference-time compute.

To quantify agreement across models, we next study cross-model consistency by correlating the task rankings induced by \method{} across LLMs.
\
\vspacesubsection{Cross-model agreement}
\label{subsec:Cross-Model_Correlation_of_Complexity}
\label{result:correlations}

A robust complexity metric should capture task-level inference difficulty that is consistent across different models, rather than behaviors specific to a particular model.

If \textbf{\method{}} is valid, different LLMs should agree on the relative complexity ranking of tasks, even if their absolute performance differs.
We benchmark our metric against three baselines to isolate this signal:
\textbf{Length:} The raw number of reasoning tokens generated (proxy for consumption). 
\textbf{Peak:} The maximum performance achievable regardless of budget (proxy for capability). 
\textbf{Budget:} The normalized token budget required to reach 90\% of peak performance (proxy for efficiency).
We quantify cross-model agreement using rank correlations: Spearman’s $\rho$ and Kendall’s $\tau_b$, computed over the global ordering of all problem--size pair instances.

Results in \mytable{tab:correlation_cross_model} yield two insights. First, \method{} captures the most stable complexity signal, achieving the highest average Kendall’s $\tau_b$ and competitive Spearman’s $\rho$, suggesting it filters out model-specific artifacts present in length-based metrics.
Second, we observe a \textbf{family clustering effect}: within-family pairs (e.g., Nemotron) show higher correlations than cross-family ones, indicating that while neural hardness is largely universal, pre-training induces subtle “dialects” of complexity that \method{} can capture.

\vspacesubsection{Prompt robustness}

\input{table/prompt_sensitivity_corr}
\label{subsec:Prompt_Sensitivity_of_Complexity}

A good metric should not only remain stable across models but also capture stable task-level differences in inference difficulty rather than superficial verbosity.
If the metric were driven primarily by verbosity, its rankings would fluctuate under different prompting strategies. Therefore, it should remain stable under prompt variations. To examine this, we evaluate five prompt variants (full prompts are provided in \mysec{appendix:prompt-sensitivity}):

\begin{itemize}[leftmargin=24pt,itemsep=1pt,topsep=0pt]
    \item \textbf{no\_step}: Remove the “step-by-step” reasoning instruction~\citep{kojima2022large}.
    \item \textbf{be\_concise}: Instruct the LLM to be concise during reasoning~\citep{renze2024benefit}.
    \item \textbf{short\_continue}: Replace the default continuation prompt with a shorter version that does not explicitly encourage self-verification.
    \item \textbf{long\_continue}: Replace the default continuation prompt with a longer version that encourages deeper reflection and rethinking.
    \item \textbf{answer\_format}: Modify the instruction to require list outputs with one item per line~\citep{long2025llmsbias}.
\end{itemize}

First, we evaluate whether the rankings induced by different prompt variants remain consistent with one another.
For each prompt setting, we report its average Spearman’s $\rho$ and Kendall’s $\tau_b$ with all other settings over 12 tasks at input length 100, averaged across five models, as shown in \mytable{tab:prompt_sensitivity_corr}; the full pairwise results are provided in \mysec{app:prompt_sensitivity_detail}.
The correlations remain consistently high across variants, indicating that \method{} is robust to prompt variations and captures stable, model-consistent differences in task complexity rather than prompt-dependent artifacts.
In contrast, other baselines—particularly the Length baseline—exhibit noticeably weaker and less reliable correlations, reflecting their greater sensitivity to prompt-induced verbosity effects.
The table in \mysec{app:prompt_sensitivity_cross_model} further shows that \method{} achieves the strongest cross-model agreement across prompt settings compared with the baselines in \mytable{tab:prompt_sensitivity_corr}.

In summary, our empirical results establish \method{} as a reliable “thermometer” for neural reasoning complexity.
It captures a signal that is both grounded in task structure and consistently recognized across models and prompt settings.
Having validated the fidelity of this signal, we next investigate its \textbf{learning utility} for optimizing the learning process.

%% file: table/correlation_cross_model.tex
\begin{table*}[t]
\centering
\setlength{\tabcolsep}{6pt}
\caption{
    \textbf{Cross-model correlation of task-level complexity rankings.}
    We report Spearman’s $\rho$ \;/\; Kendall’s $\tau_b$ between the per-task complexity ranks induced by Model A and Model B (higher is better) on 12 algorithmic problems.
    \textbf{Length}: The number of tokens in the model’s first phase CoT trace.
    \textbf{Budget}: the token budget required to reach 90\% of the model’s peak accuracy, normalized by the peak accuracy.
    \textbf{Peak}: the model’s peak accuracy.
    The results show that \textbf{\method{}} yields the best cross-model correlation among all the other heuristic complexity definitions.
}
% \vspace{-6pt}
\scalebox{0.82}{\begin{tabular}{l l l c c c c}
\toprule
\textbf{Model A} & \textbf{Model B} &  & \textbf{Length} & \textbf{Budget} & \textbf{Peak} & \textbf{\method{}} \\
\midrule

\textsc{Nemotron-49B}  & \textsc{Nemotron-253B} &
& \textbf{0.926}\;/\;\underline{0.777}
& 0.881\;/\;0.746
& 0.860\;/\;0.743
& \underline{0.919}\;/\;0.768 \\

\textsc{Nemotron-49B}  & \textsc{gpt-oss-20b}   &
& 0.711\;/\;0.518
& \textbf{0.778}\;/\;\textbf{0.609}
& 0.661\;/\;\underline{0.568}
& \underline{0.761}\;/\;0.549 \\

\textsc{Nemotron-49B}  & \textsc{gpt-oss-120b}  &
& \textbf{0.850}\;/\;\textbf{0.657}
& \underline{0.807}\;/\;0.648
& 0.716\;/\;\underline{0.655}
& 0.800\;/\;0.614 \\

\textsc{Nemotron-49B} & \textsc{Qwen-235B} &
& 0.720\;/\;0.507
& 0.650\;/\;0.511
& \underline{0.792}\;/\;\textbf{0.686}
& \textbf{0.809}\;/\;\underline{0.630} \\

\textsc{Nemotron-253B} & \textsc{gpt-oss-20b}   &
& 0.738\;/\;0.538
& \underline{0.782}\;/\;0.614
& 0.693\;/\;\underline{0.622}
& \textbf{0.837}\;/\;\textbf{0.649} \\

\textsc{Nemotron-253B} & \textsc{gpt-oss-120b}  &
& \textbf{0.864}\;/\;0.668
& 0.818\;/\;\underline{0.673}
& 0.766\;/\;0.626
& \underline{0.840}\;/\;\textbf{0.687} \\

\textsc{Nemotron-253B} & \textsc{Qwen-235B} &
& \textbf{0.838}\;/\;0.621
& 0.711\;/\;0.578
& 0.804\;/\;\textbf{0.723}
& \underline{0.810}\;/\;\underline{0.634} \\

\textsc{gpt-oss-20b}   & \textsc{gpt-oss-120b}  &
& \underline{0.872}\;/\;0.694
& 0.867\;/\;\underline{0.768}
& 0.858\;/\;0.744
& \textbf{0.913}\;/\;\textbf{0.782} \\

\textsc{gpt-oss-20b} & \textsc{Qwen-235B} &
& \textbf{0.894}\;/\;\textbf{0.721}
& 0.708\;/\;0.582
& 0.684\;/\;\underline{0.610}
& \underline{0.780}\;/\;0.606 \\

\textsc{gpt-oss-120b} & \textsc{Qwen-235B} &
& \textbf{0.948}\;/\;\textbf{0.800}
& 0.722\;/\;0.613
& 0.691\;/\;0.645
& \underline{0.874}\;/\;\underline{0.720} \\

\midrule
Average &  &
& \textbf{0.836}\;/\;0.650
& 0.772\;/\;0.634
& 0.752\;/\;\underline{0.662}
& \underline{0.834}\;/\;\textbf{0.664} \\
\bottomrule
\end{tabular}}
% \vspace{4pt}

\label{tab:correlation_cross_model}

\end{table*}

%% file: images/avg_complexity_score.tex
% Allow placement on a page with enough room and measure the wrap height automatically.
\begin{wrapfigure}{R}{0.48\linewidth}
  \vspace{-8pt}
  \centering
  \includegraphics[width=\linewidth]{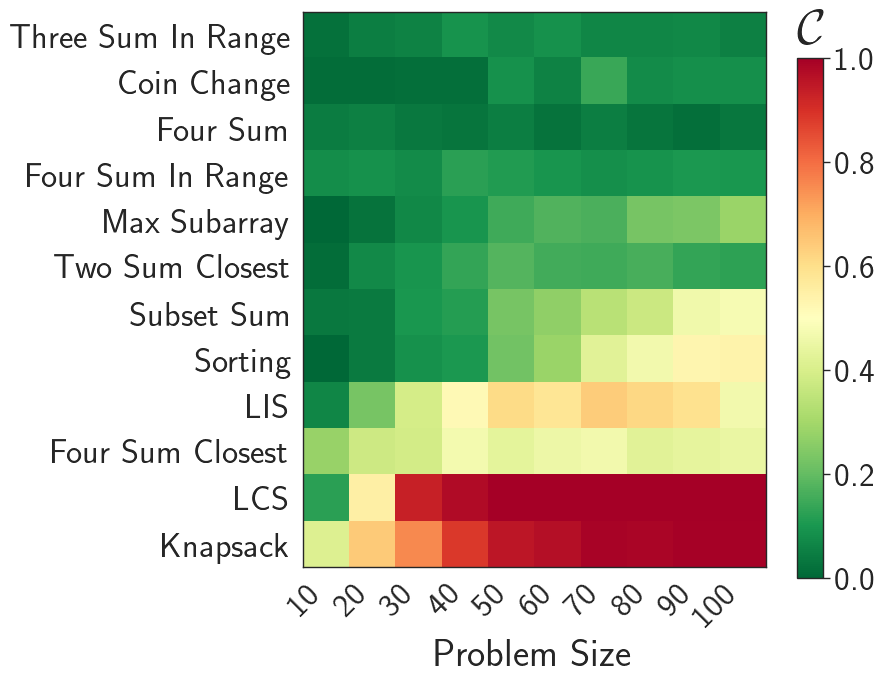}
  \caption{
  \textbf{\method{} heatmap across algorithmic tasks and problem sizes.} The heatmap shows the averaged \method{} ($\mathcal{C}$) of different algorithmic problems and corresponding problem size over 5 models (\textsc{Nemotron-49B}, \textsc{Nemotron-253B}, \textsc{Gpt-oss-20B}, \textsc{Gpt-oss-120B}, \textsc{Qwen-235B}). The results show that tasks with lower average complexity have stable scores across problem sizes, whereas tasks with higher average complexity tend to show increasing complexity as problem size increases.
  }
  \label{fig:avg_complexity_score}
  \vspace{-6pt}
\end{wrapfigure}

%% file: table/prompt_sensitivity_corr.tex
\begin{table}[t]
\centering
\small
\caption{\textbf{Prompt robustness of task-level complexity rankings.} We report the average Spearman’s $\rho$ / Kendall’s $\tau_b$ between the rankings induced by each prompt variant and all other prompt settings across five models (higher is better). \method{} achieves the highest correlations in 5 out of 6 settings and also attains the highest average correlation, indicating strong robustness to prompt variations. In particular, the \textbf{Length} baseline drops substantially under the no\_step setting, suggesting greater sensitivity to verbosity-related prompting effects.}
\label{tab:prompt_sensitivity_corr}
\scalebox{0.92}{
\begin{tabular}{@{}lccccc@{}}
\toprule
\textbf{Prompt Variants} & \textbf{Length} & \textbf{Budget} & \textbf{Peak} & \textbf{LLM Complexity} \\
\midrule
default\_setting
& 0.873 / 0.748
& \underline{0.895} / \underline{0.802}
& 0.863 / \underline{0.802}
& \textbf{0.931} / \textbf{0.849} \\

no\_step
& 0.762 / 0.634
& 0.866 / 0.787
& \underline{0.867} / \underline{0.807}
& \textbf{0.928} / \textbf{0.854} \\

be\_concise
& 0.896 / 0.791
& \underline{0.900} / \underline{0.821}
& 0.848 / 0.796
& \textbf{0.933} / \textbf{0.871} \\

short\_continue
& \underline{0.910} / 0.808
& 0.903 / \underline{0.828}
& 0.863 / 0.809
& \textbf{0.937} / \textbf{0.880} \\

long\_continue
& \underline{0.916} / 0.808
& 0.883 / 0.802
& 0.883 / \underline{0.833}
& \textbf{0.937} / \textbf{0.876} \\

answer\_format
& \textbf{0.892} / \textbf{0.774}
& 0.774 / 0.692
& 0.751 / 0.703
& \underline{0.811} / \underline{0.754} \\

\midrule
\textbf{Average}
& \underline{0.875} / 0.761
& 0.870 / 0.789
& 0.846 / \underline{0.792}
& \textbf{0.913} / \textbf{0.847} \\
\bottomrule
\end{tabular}}
\end{table}

%% file: sections/curriculum.tex
\input{table/curriculum_acc}

\vspacesection{Curriculum learning}

As \textbf{cross-model agreement} and \textbf{prompt robustness} of \method{} are shown in \mysec{sec:main_results}, a natural question arises: can this signal provide \textbf{learning utility}? Prior work suggests that easy-to-hard curricula can improve learning efficiency~\cite{bengio2009curriculum,lee2024instruction}, but their effectiveness heavily depends on the quality of the difficulty metric.
We therefore construct a curriculum based on \method{} and evaluate its effect on fine-tuning performance.

\vspacesubsection{Complexity-based curriculum sampling}
\label{curriculum:complexity-based-sampling}

Given a training instance from problem $c$ at size $n$ with normalized complexity $\mathcal{C}(c,n)\in[0,1]$, we assign it a Boltzmann-style score $s(c,n)=\exp\!\left(-\mathcal{C}(c,n)/\tau\right),$
where $\tau>0$ controls the sampling preference over difficulty levels. Smaller $\tau$ concentrates sampling on low-complexity (easy) instances, while larger $\tau$ produces a flatter distribution and increases coverage of higher-complexity (hard) instances. To avoid vanishing probabilities, we smooth the score as
$ \tilde{s}(c,n)=(1-\epsilon)s(c,n)+\frac{\epsilon}{N}, $
where $\epsilon\ge 0$ is a smoothing constant and $N$ is the number of problem--size pairs. We then normalize over all problem--size pairs:
$ P(c,n)=\frac{\tilde{s}(c,n)}{\sum_{(c',n')}\tilde{s}(c',n')}. $
Smaller $\tau$ concentrates sampling on low-complexity (easy) instances, while larger $\tau$ produces a flatter distribution and increases coverage of higher-complexity (hard) instances.

\vspacesubsection{Curriculum scheduler}
To form an easy-to-hard curriculum, we bias training toward low-complexity instances early on and gradually increase the probability of sampling more challenging instances as training progresses. Under the previous sampling scheme, this behavior can be achieved by annealing the temperature $\tau$ during the $T$ training steps, where we linearly increase $\tau$ as
$ \tau(t) \;=\; \tau_0 + \frac{t - 1}{T - 1}\,(\tau_1-\tau_0),  \qquad t\in\{1,\dots, T\}, $
where $\tau_0=\tau(1)$ denotes the initial temperature at the start of training and $\tau_1=\tau(T)$ denotes the final temperature at the end of training.

\vspacesubsection{Curriculum learning results}
We evaluate whether \method{} can provide a reliable curriculum signal for SFT. Concretely, we fine-tune \textsc{Qwen3-1.7B} on a mixture of problem--size instances $(c,n)$ sampled from the complexity-based distribution in the previous paragraph.

\noindent \textbf{Training data and supervision.}
For each training instance, we use a reasoning trace generated by \textsc{Nemotron-253B} as the supervision signal (\ie a teacher-produced chain-of-thought / ``thinking'' trace paired with the final answer). We then perform SFT on \textsc{Qwen3-1.7B} to learn to reproduce the teacher-style reasoning and answers.

\noindent \textbf{Curriculum vs.\ baselines.}
We compare data-ordering strategies under identical SFT settings (\mysec{appendix:SFT_detail}). All curricula share the same weighted sampling form $P_{\tau(t)}(\cdot)$~(\mysec{curriculum:complexity-based-sampling}) and temperature annealing schedule $\tau(t)$, differing only in the per-instance difficulty score used in the weights:
\begin{itemize}[leftmargin=24pt,itemsep=1pt,topsep=0pt]
    \item \textbf{Pre-trained} uses \textsc{Qwen3-1.7B} pre-trained weight.
    \item \textbf{Random SFT} uniformly samples instances from the training pool.
    \item \textbf{\method{} curriculum} uses $\mathcal{C}(c,n)$, \ie our \method{} score.
    \item \textbf{Length curriculum}
    uses the number of tokens in the supervision signal as difficulty.
    \item \textbf{Time curriculum}
    measures the task execution time in a Linux environment as difficulty.
    \item \textbf{Budget curriculum}: difficulty is the compute-budget proxy defined in \mysec{result:correlations}.

\end{itemize}

\noindent \textbf{Results and analysis.}

\mytable{tab:curriculum_acc} summarizes accuracy under different curricula using 3 different seeds. The SFT training set contains ten \emph{seen} algorithm families: \emph{sorting, max subarray, coin change, two sum closest, subset sum, four sum, four sum in range, LCS, knapsack}, and \emph{four sum closest}. We hold out two \emph{unseen} algorithms (\emph{LIS} and \emph{three sum in range}) to evaluate generalization beyond the training distribution. We also verify that SFT exhibits the expected seen--unseen gap and meaningful adaptation; see \mysec{appendix:effectiveness_of_SFT} for details.
Overall, the \method{} curriculum outperforms all other curricula on the \emph{seen} algorithms, indicating that complexity provides an effective signal for organizing SFT.
It also improves accuracy on the \emph{unseen} algorithms, suggesting that the learned reasoning patterns transfer beyond the training set.

%% file: table/curriculum_acc.tex
\begin{table}[b]
\centering
\small
\setlength{\tabcolsep}{8pt}
\renewcommand{\arraystretch}{1.15}
\caption{
\textbf{Accuracy usage under different curricula.}
We fine-tune \textsc{Qwen3-1.7B} with teacher-generated supervision from \textsc{Nemotron-253B} using 3 different seeds. 
All curriculum variants use the same sampling form $P \propto \exp(-\mathcal{C}/\tau)$ with an annealed temperature to induce an easy-to-hard schedule, differing only in the choice of complexity score $\mathcal{C}$: \textbf{Time} (wall-clock execution time), \textbf{Length}, \textbf{Budget} (defined in \mytable{tab:correlation_cross_model}), and \textbf{\method{}}; \textbf{Random} uses uniform sampling, and \textbf{Pre-trained} denotes the base model without SFT. 
\textbf{\method{}} achieves the highest average accuracy and lowest average token usage, supporting its effectiveness for curriculum design.
}
\label{tab:curriculum_acc}
\begin{tabular}{lcc}
\toprule
\textbf{Method} & \textbf{Seen} & \textbf{Unseen} \\
\midrule
Pre-traine & $42.75 \pm 0.00\%$ & $46.25 \pm 0.00\%$ \\
Time & $38.92 \pm 0.80\%$ & $41.25 \pm 2.17\%$ \\
Length & $44.17 \pm 1.01\%$ & $45.42 \pm 4.02\%$ \\
Budget & $43.33 \pm 0.52\%$ & $43.75 \pm 3.31\%$ \\
Random & $41.75 \pm 2.63\%$ & $41.67 \pm 1.44\%$ \\
\method{} & $\mathbf{44.42 \pm 0.76\%}$ & $\mathbf{48.33 \pm 5.05\%}$ \\
\bottomrule
\end{tabular}
\end{table}

%% file: sections/traditional_complexity.tex
\vspacesection{Fitting \method{} to computational complexity}

We investigate how \method{} relates to traditional complexity metrics by predicting its ranking from three measures: \textit{output length}, \textit{space complexity}, and \textit{time complexity}. Output length captures the expected answer length; for example, sorting outputs all $n$ elements, whereas four sum outputs only four.
For each model, we fit non-negative weights $(\alpha,\beta,\gamma)$ summing to one over the three induced rankings. We compute $s_c=\alpha r^{\text{out}}_c+\beta r^{\text{space}}_c+\gamma r^{\text{time}}_c$ and select the weights that maximize Spearman’s $\rho$ between the ranking induced by $s_c$ and the \method{} ranking.

\input{table/fitting_traditional_complexity_and_deepmath}

\mytable{tab:fitted_weights} reports the resulting mixture weights. Overall, output length consistently receives substantial weight, suggesting that it is informative for algorithmic task complexity. While weights vary across models, models within the same family exhibit similar patterns, possibly reflecting shared reasoning strategies, such as similar reliance on intermediate states. Beyond this linear decomposition, we further analyze pairwise correlations in \mysec{subsec:traditional_complexity}.

%% file: table/fitting_traditional_complexity_and_deepmath.tex
\begin{table}[t]
\centering
\small

\begin{minipage}[t]{0.48\linewidth}
\centering
\setlength{\tabcolsep}{6pt}
\caption{
\textbf{Fitted problem-level mixture weights over computational complexity measures.}
Let $r^{\text{out}}_c$, $r^{\text{space}}_c$, and $r^{\text{time}}_c$ denote rankings induced by output length, auxiliary-space complexity, and time complexity, respectively, and $r^{\text{model}}_c$ the \method{} ranking. Output length refers to the task-implied output length rather than the model’s actual response length. We grid-search nonnegative weights $(\alpha,\beta,\gamma)$ summing to one and compute $s_c=\alpha r^{\text{out}}_c+\beta r^{\text{space}}_c+\gamma r^{\text{time}}_c$, selecting the weights that maximize Spearman’s $\rho$ between the induced ranking and $r^{\text{model}}_c$. This provides a simple decomposition of \method{} across the three proxies.
}
\label{tab:fitted_weights}
\scalebox{0.85}{
\begin{tabular}{lccc}
\toprule
\textbf{Model} & \textbf{$\alpha$ (Output)} & \textbf{$\beta$ (Space)} & \textbf{$\gamma$ (Time)} \\
\midrule
\textsc{Nemotron-49B}   & 0.49 & 0.51 & 0.00 \\
\textsc{Nemotron-253B}  & 0.48 & 0.51 & 0.01 \\
\textsc{gpt-oss-20b}    & 0.55 & 0.25 & 0.20 \\
\textsc{gpt-oss-120b}   & 0.58 & 0.23 & 0.19 \\
\textsc{Qwen-235B}      & 0.56 & 0.25 & 0.19 \\
\bottomrule
\end{tabular}
}
\end{minipage}
\hfill
\begin{minipage}[t]{0.48\linewidth}
\centering
\setlength{\tabcolsep}{6pt}
\renewcommand{\arraystretch}{1.15}
\caption{
\textbf{Task-level complexity correlation on DeepMath.}
We report cross-model correlation (averaged per model over pairwise comparisons) and correlation with DeepMath difficulty annotations using Spearman’s $\rho$ and Kendall’s $\tau_b$. \method{} shows substantial cross-model consistency but near-zero correlation with human-annotated difficulty, suggesting that it captures a complementary dimension of LLM-specific problem complexity. Full pairwise correlations are provided in \mysec{app:additional_deepmath_results}.
}
\label{tab:deepmath_instance_correlation}
\scalebox{0.85}{
\begin{tabular}{l cc}
\toprule
\textbf{Model} & \textbf{Cross-model} & \textbf{vs.\ DeepMath} \\
\midrule
\textsc{Nemotron-49B}  & 0.544 / 0.482 & -0.063 / -0.051 \\
\textsc{Nemotron-253B} & 0.500 / 0.451 & -0.071 / -0.059 \\
\textsc{gpt-oss-20b}   & 0.460 / 0.423 &  0.001 /  0.001 \\
\textsc{gpt-oss-120b}  & 0.478 / 0.441 & -0.002 / -0.002 \\
\textsc{Qwen-235B}     & 0.546 / 0.491 & -0.111 / -0.093 \\
\bottomrule
\end{tabular}
}
\end{minipage}

\end{table}

%% file: sections/real_world.tex
\input{table/rule_arena_results}
\vspacesection{Generalizing to real-world problems}

To examine if our \method{} 
goes beyond synthetic algorithmic problems and 
generalizes to real-world tasks,
we conduct experiments on DeepMath~\cite{he2025deepmath} and RuleArena~\citep{zhou2024rulearena}, which were released after the knowledge cutoff of all evaluated models (\ie no data contamination issue).

\noindent \textbf{DeepMath.}
DeepMath~\cite{he2025deepmath} contains math problems with difficulty labels assigned by GPT-4o~\citep{hurst2024gpt4o} using Art of Problem Solving guidelines~\citep{aops_competition_ratings} to approximate human-perceived difficulty.
We evaluate 200 problems, comparing \method{} rankings across models and against these difficulty labels.
As shown in~\mytable{tab:deepmath_instance_correlation}, \method{} rankings exhibit substantial cross-model agreement, consistent with~\mysec{sec:main_results}, but near-zero correlation with DeepMath’s labels.
This suggests that human-oriented mathematical difficulty and LLM inference-time complexity capture different aspects of problem difficulty.

\noindent \textbf{RuleArena.}
RuleArena~\citep{zhou2024rulearena} contains real-world reasoning tasks across three categories—airline baggage fees, NBA transactions, and tax regulations—with three difficulty levels each. We treat each category–difficulty pair as a task and evaluate cross-model agreement over the resulting nine tasks. As shown in \mytable{tab:correlation_cross_model_rulearena}, \method{} achieves the highest average rank correlation across model pairs and ranks within the top two in most comparisons, demonstrating strong cross-model consistency on real-world reasoning tasks.

%% file: table/rule_arena_results.tex
\begin{table*}[t]
\centering
\small
\setlength{\tabcolsep}{6pt}
\caption{
\textbf{Cross-model consistency of task-level complexity rankings on RuleArena.}
We report Spearman’s $\rho$ \;/\; Kendall’s $\tau_b$ between the rankings of the nine RuleArena tasks induced by Model A and Model B (higher is better). 
\method{} achieves the strongest cross-model agreement.
}
\scalebox{0.85}{\begin{tabular}{l l l c c c c}
\toprule
\textbf{Model A} & \textbf{Model B} &  & \textbf{Length} & \textbf{Budget} & \textbf{Peak} & \textbf{\method{}} \\
\midrule
\textsc{Nemotron-49B}  & \textsc{Nemotron-253B} &
& \underline{0.983}\;/\;\underline{0.944}
& \underline{0.983}\;/\;0.939
& 0.897\;/\;0.818
& \textbf{1.000}\;/\;\textbf{1.000} \\

\textsc{Nemotron-49B}  & \textsc{gpt-oss-20b}   &
& 0.817\;/\;0.722
& 0.868\;/\;0.765
& \textbf{0.945}\;/\;\textbf{0.883}
& \underline{0.881}\;/\;\underline{0.783} \\

\textsc{Nemotron-49B}  & \textsc{gpt-oss-120b}  &
& \underline{0.633}\;/\;\underline{0.556}
& \textbf{0.966}\;/\;\textbf{0.899}
& 0.573\;/\;0.418
& \textbf{0.966}\;/\;\textbf{0.899} \\

\textsc{Nemotron-49B}  & \textsc{Qwen-235B} &
& 0.800\;/\;0.611
& 0.828\;/\;\underline{0.758}
& \underline{0.879}\;/\;\underline{0.758}
& \textbf{1.000}\;/\;\textbf{1.000} \\

\textsc{Nemotron-253B} & \textsc{gpt-oss-20b}   &
& 0.833\;/\;0.778
& \textbf{0.928}\;/\;\textbf{0.824}
& 0.843\;/\;0.706
& \underline{0.881}\;/\;\underline{0.783} \\

\textsc{Nemotron-253B} & \textsc{gpt-oss-120b}  &
& 0.617\;/\;0.500
& \underline{0.932}\;/\;\underline{0.841}
& 0.735\;/\;0.597
& \textbf{0.966}\;/\;\textbf{0.899} \\

\textsc{Nemotron-253B} & \textsc{Qwen-235B} &
& 0.750\;/\;0.556
& 0.897\;/\;0.818
& \underline{0.983}\;/\;\underline{0.939}
& \textbf{1.000}\;/\;\textbf{1.000} \\

\textsc{gpt-oss-20b}   & \textsc{gpt-oss-120b}  &
& 0.683\;/\;0.500
& \underline{0.862}\;/\;\textbf{0.761}
& 0.498\;/\;0.377
& \textbf{0.867}\;/\;\underline{0.722} \\

\textsc{gpt-oss-20b} & \textsc{Qwen-235B} &
& 0.383\;/\;0.333
& \textbf{0.894}\;/\;\underline{0.765}
& 0.826\;/\;0.647
& \underline{0.881}\;/\;\textbf{0.783} \\

\textsc{gpt-oss-120b} & \textsc{Qwen-235B} &
& 0.467\;/\;0.389
& \underline{0.729}\;/\;\underline{0.667}
& 0.684\;/\;0.537
& \textbf{0.966}\;/\;\textbf{0.899} \\

\midrule
Average &  &
& 0.697\;/\;0.589
& \underline{0.889}\;/\;\underline{0.804}
& 0.786\;/\;0.668
& \textbf{0.941}\;/\;\textbf{0.877} \\
\bottomrule
\end{tabular}}

\label{tab:correlation_cross_model_rulearena}
\end{table*}

%% file: sections/conclusion.tex
\vspacesection{Conclusion}

We propose \method{}, a metric that quantifies problem complexity for LLMs via the area over the accuracy–token curve.
We show that \method{} captures stable task-level differences in inference difficulty, yielding rankings that are highly consistent across diverse model families.
We further demonstrate its practical utility by constructing easy-to-hard curricula for supervised fine-tuning that improve accuracy while reducing token consumption. Finally, we show that \method{} generalizes to real-world tasks, producing more stable cross-model correlations than human-annotated difficulty labels. Overall, our findings highlight the importance of defining complexity in terms of how LLMs allocate inference-time compute, rather than relying solely on classical algorithmic hierarchies.

%% file: sections/appendix/main.tex
\clearpage

\input{sections/appendix/task_table}
\input{sections/appendix/experiment_detail}
\input{sections/appendix/additional_results}
\input{sections/appendix/correlation_traditional}
\input{sections/appendix/SFT_detail}

%% file: sections/appendix/task_table.tex
\section{Problem table}

\label{appendix:task_table}

\mytable{tab:task-complexity} provides a complete specification of the evaluation suite. For each problem, we list the problem name and include a representative instance prompt drawn from the same template used during evaluation, illustrating the exact input format (\eg list-structured inputs or paired strings) and the required output type (\eg an ordered list, a subset, a subsequence, or a scalar objective value). All problems are classical algorithmic problems with verifiable outputs, enabling consistent automatic checking across instances. As described in our data construction procedure, we instantiate each problem at multiple problem sizes (from 10 to 100 in increments of 10), so Table 8 serves as a concrete reference for how each problem is presented to the model while the evaluation varies the underlying problem scale. \mytable{tab:complexity_table} reports the annotated worst-case time and auxiliary-space complexity of the 12 tasks. We use these annotations to form the task-level time and space complexity rankings used in our experiments.

%% file: sections/appendix/experiment_detail.tex
\section{Prompt details}

\subsection{Prompt used for inference}

Our inference pipeline consists of two stages: (1) long-chain generation and (2) answer generation from a truncated prefix.

\textbf{Long-chain generation.} To ask the LLM generate the long chain, we use the following prompt:

\begin{mdframed}[nobreak=true]
    \texttt{\{Problem Specific Prompt\} Provide your reasoning step by step without using external programming functions. Lastly, present the final answer in the exact format: Answer: [...]}
\end{mdframed}

\textbf{Answer generation from a truncated prefix.} To ask the LLM to generate an answer based on the truncated prefix, we use the following prompt:

\begin{mdframed}[nobreak=true]
    \texttt{\{Problem Specific Prompt\} Compute the result manually without using external programming functions. Present the final answer in the exact format: Answer: [...]}
\end{mdframed}

The \texttt{\{Problem Specific Prompt\}} varies across problems. For example, for \emph{sorting}, one instance is: 

\begin{mdframed}[nobreak=true]
\texttt{Given the list [36, 23, 34, 73, 98], sort the list in ascending order.}
\end{mdframed}

We provide a sample prompt for each task in \myapp{appendix:task_table}.

\subsection{Continuation prompt}
\label{appendix:continue-prompt}

If the LLM terminates generation before reaching the maximum token budget, we append the following prompt to encourage it to continue reasoning until the budget is exhausted.

\begin{mdframed}[nobreak=true]
    \texttt{Please review your previous reasoning step by step and check for any errors, incorrect assumptions, or logical inconsistencies. If issues are found, revise the reasoning accordingly.}
\end{mdframed}

\subsection{Detailed prompts for prompt sensitivity experiments}
\label{appendix:prompt-sensitivity}

In \mysec{subsec:Prompt_Sensitivity_of_Complexity}, we show that \method{} is robust to prompt variations. Here, we detail the prompt variants used in our experiments.

\begin{itemize}
     \item \textbf{no\_step}: For long-chain generation, we replace the default prompt with:

    \begin{mdframed}[nobreak=true]
        \texttt{Compute the result manually without using external programming functions. Do not provide any reasoning or explanation. Lastly, present the final answer in the exact format: Answer: ...}
    \end{mdframed}
    
    \item \textbf{be\_concise}: For long-chain generation, we use the following prompt:

    \begin{mdframed}[nobreak=true]
        \texttt{Provide your reasoning step by step without using external programming functions. Lastly, present the final answer in the exact format: Answer: [...] Be concise.}
    \end{mdframed}
    
    \item \textbf{short\_continue}: We replace the default continuation prompt with:
    
    \begin{mdframed}[nobreak=true]
        \texttt{Please continue thinking.}
    \end{mdframed}
    
    \item \textbf{long\_continue}: We replace the default continuation prompt with:

    \begin{mdframed}[nobreak=true]
        \texttt{You have previously generated a detailed reasoning process for this problem. Now, carefully review your entire reasoning step by step. For each step, check whether the logic is valid, whether any assumptions are incorrect, and whether any intermediate calculations or conclusions are flawed. If you identify any mistakes, inconsistencies, or gaps in reasoning, revise the reasoning accordingly and ensure that the corrected version is logically coherent and consistent with the problem statement. After completing this verification and correction process, provide the final answer. Ensure that the final answer is correct and consistent with your revised reasoning.}
    \end{mdframed}

    \item \textbf{answer\_format}: For answer generation, we replace the default prompt with:

    \begin{mdframed}[nobreak=true]
    \texttt{Compute the result manually without using external programming functions. generate your list such that each element is in a new line, and present the final answer in the exact format:\textbackslash n\textbackslash nAnswer:\textbackslash na\textbackslash nb\textbackslash nc\textbackslash nd}
    \end{mdframed}
    
\end{itemize}

\input{table/task_table}
\input{table/complexity_table}

%% file: table/task_table.tex
\begin{table}[H]
\centering
\scriptsize
\setlength{\tabcolsep}{3pt}
\renewcommand{\arraystretch}{1.15}
\caption{Tasks used in our evaluation.}
\begin{adjustbox}{max width=0.92\linewidth,center}
\begin{tabularx}{0.92\linewidth}{l X}
\toprule
Task & Sample Prompt \\
\midrule
Max Subarray &
Given the list nums = [-38, 41, 24, -42, 46], find the maximum possible sum of a contiguous subarray. \\
\midrule

Sorting &
Given the list [36, 23, 34, 73, 98], sort the list in ascending order. \\
\midrule

Two Sum Closest &
Given a list of integers of length 5: [25, 182, 149, 17, 193], find two numbers whose sum is closest to the target value 31. If multiple answers exist, return any one of them. \\
\midrule

Longest Increasing Subsequence (LIS) &
Given the list [5, 4, 3, 1, 2], find the longest increasing subsequence. \\
\midrule

Three Sum (In Range) &
Given a list of integers of length 5: [118, 46, 52, 110, 107], find any 3 numbers whose sum is in the range [268, 276] (inclusive). \\
\midrule

Four Sum &
Given the list of integers of length 5: [96, 58, 60, 56, 50], find any four numbers that sum to 264. \\
\midrule

Four Sum (In Range) &
Given the list of integers of length 5: [107, 20, 60, 99, 146], find any 4 numbers whose sum is in the range [282, 287] (inclusive). \\
\midrule

Four Sum Closest &
Given the list of integers of length 5: [131, 19, 173, 12, 187], find four numbers whose sum is closest to the target value 449. There is no exact solution where the sum equals 449. If multiple answers achieve the same closest distance, return any one of them. \\
\midrule

Coin Change &
Given coin denominations coins = [1, 5, 9, 11, 14, 18, 25, 26, 27] (unlimited supply of each), find the minimum number of coins needed to make exactly amount = 11. Return the minimum number of coins as an integer. \\
\midrule

Subset Sum &
Given the list [96, 11, 82, 59, 54] and the target sum 302, find a subset of the list that sums exactly to the target. \\
\midrule

Longest Common Subsequence (LCS) &
Given two strings s1 and s2, find a longest common subsequence. s1 = ejbpc s2 = kefiq Return any one LCS string (if multiple exist, any is acceptable). \\
\midrule

Knapsack (0/1) &
You are given n items. items = [(weight:18,value:3), (weight:11,value:7), (weight:4,value:17), (weight:18,value:8), (weight:15,value:17)] Capacity W = 26 Choose a subset of items (each item at most once) to maximize total value without exceeding capacity. \\
\bottomrule
\end{tabularx}
\end{adjustbox}

\label{tab:task-complexity}
\end{table}

%% file: table/complexity_table.tex
\begin{table}[H]
\centering
\caption{\textbf{Annotated time and auxiliary-space complexity for the 12 algorithmic tasks.}
Time bounds follow the tightest widely accepted \emph{worst-case} complexity under the standard computation model. Space reports worst-case \emph{auxiliary} memory, excluding input/output storage.
Here $n$ is input list length; for Longest Common Subsequence, $n,m$ are the two sequence lengths; for Coin Change, $A$ is the target amount and $k$ is the number of coin types; for Knapsack, $W$ is capacity; for Subset Sum, $T$ is the target sum.
$^{\dagger}$Assuming in-place sorting; recursion stack is $O(\log n)$ if applicable.
$^{\ddagger}$Subset Sum is NP-complete in $n$; we annotate the exponential worst-case in $n$ while noting its pseudo-polynomial DP dependence on $T$.}
\label{tab:complexity_table}

\vspace{2pt}

\small
\setlength{\tabcolsep}{4pt}
\renewcommand{\arraystretch}{1.08}
\begin{tabularx}{\linewidth}{lccX}
\toprule
\textbf{Task} & \textbf{Time} & \textbf{Aux. Space} & \textbf{Notes / Parameters} \\
\midrule
Max Subarray & $O(n)$ & $O(1)$ & Kadane scan \\
Sorting & $O(n\log n)$ & $O(1)^{\dagger}$ & Comparison-based sorting \\
Two Sum Closest & $O(n\log n)$ & $O(1)^{\dagger}$ & Sort + two pointers \\
LIS & $O(n\log n)$ & $O(n)$ & Patience sorting \\
Three Sum (In Range) & $O(n^{2})$ & $O(1)^{\dagger}$ & Sort + two pointers \\
Four Sum & $O(n^{3})$ & $O(1)^{\dagger}$ & Sort + 2 loops + two pointers \\
Four Sum (In Range) & $O(n^{3})$ & $O(1)^{\dagger}$ & Same as Four Sum \\
Four Sum Closest & $O(n^{3})$ & $O(1)^{\dagger}$ & Same as Four Sum \\
Coin Change & $O(kA)$ & $O(A)$ & DP over amount $A$; $k$ coin types \\
LCS & $O(nm)$ & $O(\min(n,m))$ & DP time; Hirschberg space \\
Knapsack (0/1) & $O(nW)$ & $O(W)$ & DP over capacity $W$ \\
Subset Sum & $O(2^n)^{\ddagger}$ & $O(T)^{\ddagger}$ & NP-complete; pseudo-polynomial DP in target $T$ \\
\bottomrule
\end{tabularx}
\end{table}

%% file: sections/appendix/additional_results.tex
\input{table/detailed_output_token}
\input{table/pairwise_prompt_sens}
\input{table/prompt_sensitivity_cross_model}
\input{table/deepmath_detail}

\section{Additional Results}
\subsection{Task-level average output length}
\label{app:avg_outlen}

For each task, we measure the length of the model’s generated output in tokens and report the task-level average. Concretely, for every problem, we aggregate token counts across all evaluated instances and problem sizes under the same decoding configuration. We report the results in \mytable{tab:avg_out_tokens_by_task}.

\subsection{Additional prompt sensitivity results}

Here, we provide additional results from the prompt sensitivity experiments in \mysec{subsec:Prompt_Sensitivity_of_Complexity}, further demonstrating that \method{} is robust to prompt variations.
Specifically, \mysec{app:prompt_sensitivity_detail} presents the full pairwise correlations between different prompt settings, providing a more fine-grained view of prompt-level agreement.
We then analyze cross-model consistency under each prompt variant in \mysec{app:prompt_sensitivity_cross_model}, showing that \method{} maintains strong agreement across diverse models even under substantial prompt changes.

\subsubsection{Pairwise prompt sensitivity results}

\label{app:prompt_sensitivity_detail}

To complement the aggregate prompt-sensitivity results in \mytable{tab:prompt_sensitivity_corr}, we provide the full pairwise correlations between all prompt settings in \mytable{tab:prompt_sensitivity_pairwise}.

For each prompt setting, we report its Spearman’s $\rho$ and Kendall’s $\tau_b$ with every other setting over 12 tasks at input length 100, averaged across five models.

The Average row within each block corresponds to the aggregate value reported in \mytable{tab:prompt_sensitivity_corr}, while the Overall Average row summarizes agreement across all prompt pairs.

\subsubsection{Cross-model consistency under prompt settings}

\label{app:prompt_sensitivity_cross_model}

We further evaluate cross-model consistency under each prompt variant by averaging Spearman's $\rho$ and Kendall's $\tau_b$ across all model pairs.
As shown in \mytable{tab:prompt_sensitivity_cross_model}, \method{} achieves the highest overall agreement among all metrics, with an average correlation of $0.903$ / $0.780$.
It ranks first in three out of five settings and second in the remaining two, while the \textbf{Length} baseline drops notably under \textbf{no\_step}, indicating higher sensitivity to prompt-induced verbosity changes.

\subsection{Additional deepmath results}
\label{app:additional_deepmath_results}

We report the full pairwise cross-model correlations of task-level complexity rankings on DeepMath in \mytable{tab:deepmath_cross_model_detail}. For each pair of models, we compute Spearman’s $\rho$ and Kendall’s $\tau_b$ over instance-level complexity scores induced by \method{}. These results complement the aggregated statistics in the main text by providing a more fine-grained view of agreement across model families.

Overall, we observe moderate to strong positive correlations across most model pairs, indicating that \method{} induces relatively consistent complexity rankings even across models with different architectures and scales.

Notably, models within the same family (e.g., \textsc{Nemotron} or \textsc{Gpt-oss}) tend to exhibit higher agreement, while cross-family correlations are slightly lower but remain positive. This pattern suggests that while some variation exists due to model-specific behaviors, the proposed metric captures a shared notion of inference-time difficulty.

% [original]
% \subsection{Curriculum Learning results with standard deviation}
% Here, we report the standard deviation of the accuracy and token.
% \input{table/acc_std}
% \input{table/tok_std}

% [revised]
% Tables 14 and 15 are omitted from the paper.

%% file: table/detailed_output_token.tex
\begin{table}
  \centering
  \caption{Average number of output tokens for each problem under step-by-step prompting, aggregated across all problem sizes and LLMs, sorted from low to high.}
  \label{tab:avg_out_tokens_by_task}
  \begin{tabular}{lr}
    \toprule
    Task & Avg. output tokens \\
    \midrule
    Four Sum                    & 1968 \\
    Four Sum In Range           & 2851 \\
    Three Sum In Range          & 2854 \\
    Coin Change                 & 3404 \\
    Subset Sum                  & 3508 \\
    Two Sum Closest             & 4094 \\
    Max Subarray                & 4411 \\
    Sorting                     & 5466 \\
    LIS                         & 7907 \\
    LCS                         & 10505 \\
    Four Sum Closest            & 10926 \\
    Knapsack                    & 12421 \\
    \bottomrule
  \end{tabular}
\end{table}

%% file: table/pairwise_prompt_sens.tex
\begin{table*}[t]

\centering

\scriptsize

\setlength{\tabcolsep}{4pt}

\renewcommand{\arraystretch}{1.12}

\caption{
\textbf{Detailed pairwise prompt sensitivity results.}
For each prompt setting, we report its pairwise Spearman's $\rho$ \;/\; Kendall's $\tau_b$ with every other setting over 12 tasks at input length 100, averaged across five models.
The Average row within each block summarizes the mean agreement of that prompt setting with all others, while Overall Average averages across all prompt-setting pairs.
Bold and underline denote the highest and second-highest values within each row, with Spearman's $\rho$ and Kendall's $\tau_b$ ranked independently.
}

\label{tab:prompt_sensitivity_pairwise}

\resizebox{\textwidth}{!}{

\begin{tabular}{llcccc}

\toprule

\textbf{Prompt Setting} & \textbf{Compared Setting} & \textbf{Length} & \textbf{Budget} & \textbf{Peak} & \textbf{LLM Complexity} \\

\midrule

\multirow{6}{*}{default\_setting}
& no\_step
& 0.751 / 0.612
& 0.887 / 0.783
& \underline{0.893} / \underline{0.828}
& \textbf{0.940} / \textbf{0.848} \\
& be\_concise
& 0.885 / 0.763
& \underline{0.904} / \underline{0.805}
& 0.853 / 0.779
& \textbf{0.943} / \textbf{0.855} \\
& short\_continue
& 0.919 / 0.806
& \underline{0.934} / \underline{0.867}
& 0.871 / 0.800
& \textbf{0.956} / \textbf{0.884} \\
& long\_continue
& 0.929 / 0.812
& 0.899 / 0.803
& \underline{0.943} / \textbf{0.898}
& \textbf{0.962} / \underline{0.885} \\
& answer\_format
& \textbf{0.881} / 0.745
& 0.850 / \underline{0.754}
& 0.754 / 0.704
& \underline{0.854} / \textbf{0.775} \\
& \textbf{Average}
& 0.873 / 0.748
& \underline{0.895} / \underline{0.802}
& 0.863 / 0.802
& \textbf{0.931} / \textbf{0.849} \\

\midrule
\multirow{6}{*}{no\_step}
& default\_setting
& 0.751 / 0.612
& 0.887 / 0.783
& \underline{0.893} / \underline{0.828}
& \textbf{0.940} / \textbf{0.848} \\
& be\_concise
& 0.747 / 0.630
& \underline{0.907} / \underline{0.838}
& 0.899 / 0.834
& \textbf{0.970} / \textbf{0.895} \\
& short\_continue
& 0.775 / 0.648
& \underline{0.921} / \underline{0.847}
& 0.895 / 0.840
& \textbf{0.967} / \textbf{0.901} \\
& long\_continue
& 0.787 / 0.667
& 0.903 / 0.832
& \underline{0.917} / \underline{0.857}
& \textbf{0.966} / \textbf{0.895} \\
& answer\_format
& \underline{0.749} / 0.612
& 0.712 / 0.634
& 0.731 / \underline{0.677}
& \textbf{0.796} / \textbf{0.728} \\
& \textbf{Average}
& 0.762 / 0.634
& 0.866 / 0.787
& \underline{0.867} / \underline{0.807}
& \textbf{0.928} / \textbf{0.854} \\

\midrule
\multirow{6}{*}{be\_concise}
& default\_setting
& 0.885 / 0.763
& \underline{0.904} / \underline{0.805}
& 0.853 / 0.779
& \textbf{0.943} / \textbf{0.855} \\
& no\_step
& 0.747 / 0.630
& \underline{0.907} / \underline{0.838}
& 0.899 / 0.834
& \textbf{0.970} / \textbf{0.895} \\
& short\_continue
& 0.945 / 0.860
& \underline{0.958} / \underline{0.886}
& 0.863 / 0.823
& \textbf{0.983} / \textbf{0.936} \\
& long\_continue
& \underline{0.966} / \underline{0.879}
& 0.944 / 0.863
& 0.898 / 0.861
& \textbf{0.972} / \textbf{0.918} \\
& answer\_format
& \textbf{0.937} / \textbf{0.824}
& 0.786 / 0.712
& 0.727 / 0.683
& \underline{0.798} / \underline{0.750} \\
& \textbf{Average}
& 0.896 / 0.791
& \underline{0.900} / \underline{0.821}
& 0.848 / 0.796
& \textbf{0.933} / \textbf{0.871} \\

\midrule
\multirow{6}{*}{short\_continue}
& default\_setting
& 0.919 / 0.806
& \underline{0.934} / \underline{0.867}
& 0.871 / 0.800
& \textbf{0.956} / \textbf{0.884} \\
& no\_step
& 0.775 / 0.648
& \underline{0.921} / \underline{0.847}
& 0.895 / 0.840
& \textbf{0.967} / \textbf{0.901} \\
& be\_concise
& 0.945 / 0.860
& \underline{0.958} / \underline{0.886}
& 0.863 / 0.823
& \textbf{0.983} / \textbf{0.936} \\
& long\_continue
& \underline{0.958} / \underline{0.860}
& 0.926 / 0.845
& 0.898 / 0.839
& \textbf{0.977} / \textbf{0.921} \\
& answer\_format
& \textbf{0.954} / \textbf{0.867}
& 0.775 / 0.694
& 0.788 / 0.741
& \underline{0.799} / \underline{0.759} \\
& \textbf{Average}
& \underline{0.910} / 0.808
& 0.903 / \underline{0.828}
& 0.863 / 0.809
& \textbf{0.937} / \textbf{0.880} \\

\midrule
\multirow{6}{*}{long\_continue}
& default\_setting
& 0.929 / 0.812
& 0.899 / 0.803
& \underline{0.943} / \textbf{0.898}
& \textbf{0.962} / \underline{0.885} \\
& no\_step
& 0.787 / 0.667
& 0.903 / 0.832
& \underline{0.917} / \underline{0.857}
& \textbf{0.966} / \textbf{0.895} \\
& be\_concise
& \underline{0.966} / \underline{0.879}
& 0.944 / 0.863
& 0.898 / 0.861
& \textbf{0.972} / \textbf{0.918} \\
& short\_continue
& \underline{0.958} / \underline{0.860}
& 0.926 / 0.845
& 0.898 / 0.839
& \textbf{0.977} / \textbf{0.921} \\
& answer\_format
& \textbf{0.938} / \textbf{0.824}
& 0.745 / 0.665
& 0.758 / 0.710
& \underline{0.807} / \underline{0.758} \\
& \textbf{Average}
& \underline{0.916} / 0.808
& 0.883 / 0.802
& 0.883 / \underline{0.833}
& \textbf{0.937} / \textbf{0.876} \\

\midrule
\multirow{6}{*}{answer\_format}
& default\_setting
& \textbf{0.881} / 0.745
& 0.850 / \underline{0.754}
& 0.754 / 0.704
& \underline{0.854} / \textbf{0.775} \\
& no\_step
& \underline{0.749} / 0.612
& 0.712 / 0.634
& 0.731 / \underline{0.677}
& \textbf{0.796} / \textbf{0.728} \\
& be\_concise
& \textbf{0.937} / \textbf{0.824}
& 0.786 / 0.712
& 0.727 / 0.683
& \underline{0.798} / \underline{0.750} \\
& short\_continue
& \textbf{0.954} / \textbf{0.867}
& 0.775 / 0.694
& 0.788 / 0.741
& \underline{0.799} / \underline{0.759} \\
& long\_continue
& \textbf{0.938} / \textbf{0.824}
& 0.745 / 0.665
& 0.758 / 0.710
& \underline{0.807} / \underline{0.758} \\
& \textbf{Average}
& \textbf{0.892} / \textbf{0.774}
& 0.774 / 0.692
& 0.751 / 0.703
& \underline{0.811} / \underline{0.754} \\

\midrule
\textbf{Overall Average} &

& \underline{0.875} / 0.761

& 0.870 / 0.789

& 0.846 / \underline{0.792}

& \textbf{0.913} / \textbf{0.847} \\
\bottomrule

\end{tabular}

}

\end{table*}

%% file: table/prompt_sensitivity_cross_model.tex
\begin{table}[t]
\centering
\small
\caption{The average Spearman’s $\rho$ \;/\; Kendall’s $\tau_b$ of complexity rankings across all model pairs under each prompt variant (higher is better).
\textbf{LLM Complexity} achieves the strongest overall cross-model agreement, with the highest average correlation across prompt variations and consistently high performance across settings.}
\label{tab:prompt_sensitivity_cross_model}
\begin{tabular}{@{}lcccc@{}}
\toprule
\textbf{Ablation} & \textbf{Length} & \textbf{Budget} & \textbf{Peak} & \textbf{LLM Complexity} \\
\midrule
no\_step
& 0.618 / 0.491
& \underline{0.814} / \underline{0.681}
& 0.732 / 0.651
& \textbf{0.905} / \textbf{0.785} \\

be\_concise
& \underline{0.888} / 0.758
& 0.887 / \underline{0.763}
& 0.725 / 0.643
& \textbf{0.899} / \textbf{0.776} \\

short\_continue
& \textbf{0.923} / \textbf{0.806}
& 0.868 / 0.738
& 0.803 / 0.738
& \underline{0.891} / \underline{0.765} \\

long\_continue
& \textbf{0.917} / \textbf{0.800}
& 0.852 / 0.728
& 0.752 / 0.663
& \underline{0.912} / \underline{0.794} \\

answer\_format
& 0.878 / 0.732
& \underline{0.896} / \textbf{0.784}
& 0.716 / 0.647
& \textbf{0.906} / \underline{0.780} \\

\midrule

\textbf{Average}
& 0.845 / 0.717
& \underline{0.863} / \underline{0.739}
& 0.746 / 0.668
& \textbf{0.903} / \textbf{0.780} \\

\bottomrule
\end{tabular}
\end{table}

%% file: table/deepmath_detail.tex
\begin{table}[t]

\centering

\small

\caption{
\textbf{Full pairwise cross-model correlation on DeepMath.}
We report Spearman’s $\rho$ and Kendall’s $\tau_b$ between task-level complexity rankings induced by \method{} for all model pairs.
Each entry is computed over instance-level complexity scores and reflects the agreement in relative task difficulty between two models.
Higher values indicate stronger alignment.
Consistent with the main results, correlations are generally moderate to strong, with higher agreement observed within model families and slightly lower (but still positive) correlations across different families.
}

\label{tab:deepmath_cross_model_detail}

\setlength{\tabcolsep}{6pt}

\renewcommand{\arraystretch}{1.15}

\resizebox{0.68\textwidth}{!}{

\begin{tabular}{l l c c}

\toprule

Model A & Model B & Spearman $\rho$ & Kendall $\tau_b$ \\

\midrule

\textsc{Nemotron-49B}  & \textsc{Nemotron-253B} & 0.679 & 0.604 \\

\textsc{Nemotron-49B}  & \textsc{Gpt-oss-20B}   & 0.419 & 0.375 \\

\textsc{Nemotron-49B}  & \textsc{Gpt-oss-120B}  & 0.421 & 0.376 \\

\textsc{Nemotron-49B}  & \textsc{Qwen-235B}     & 0.655 & 0.574 \\

\textsc{Nemotron-253B} & \textsc{Gpt-oss-20B}   & 0.357 & 0.324 \\

\textsc{Nemotron-253B} & \textsc{Gpt-oss-120B}  & 0.346 & 0.315 \\

\textsc{Nemotron-253B} & \textsc{Qwen-235B}     & 0.617 & 0.560 \\

\textsc{Gpt-oss-20B}   & \textsc{Gpt-oss-120B}  & 0.648 & 0.619 \\

\textsc{Gpt-oss-20B}   & \textsc{Qwen-235B}     & 0.414 & 0.376 \\

\textsc{Gpt-oss-120B}  & \textsc{Qwen-235B}     & 0.498 & 0.454 \\

\bottomrule

\end{tabular}

}

\end{table}

%% file: sections/appendix/correlation_traditional.tex
\input{table/grounding_traditional_complexity}

\section{The correlation of \method{} and computational complexities}

\label{subsec:traditional_complexity}

To better understand the characteristics of \method{}, we compare \method{} against two classical measures—space complexity and time complexity—and an additional proxy, \emph{output length}, defined as the expected token length of the ground-truth answer.

For each problem, we construct problem-level rankings induced by these three measures and measure their rank consistency with each model’s problem-level \method{} ranking using Spearman’s $\rho$ and Kendall’s $\tau_b$.
For each model, we first aggregate complexity scores across input lengths to obtain a single problem-level \method{} score per problem, and apply the same aggregation procedure to each classic measure to obtain problem-level rankings. We measure space complexity and time complexity using worst-case auxiliary memory usage and worst-case runtime, respectively.

\mytable{tab:grounding_traditional_complexity} reports the results.
Overall, output length shows the strongest alignment with \method{} ranking on average.
One plausible explanation is that, in our benchmark, longer gold answers typically require producing more symbols and maintaining more intermediate information. For example, in sorting, the gold answer is an ordered list of $n$ elements, so the output length grows with $n$. Generating this list autoregressively is more than “printing” $n$ tokens: at each step, the model must maintain the partially sorted prefix and choose the next element to remain consistent with both the remaining inputs and the prefix, leading to more token usage.
We also observe that the gpt-oss models correlate more strongly with space and time complexity than the Nemotron models, suggesting that gpt-oss complexity is more tightly coupled to problems that demand a larger maintained state or longer sequential computation. 

%% file: table/grounding_traditional_complexity.tex
\begin{table}[H]
\centering
\setlength{\tabcolsep}{4pt} 
\small
\caption{
\textbf{Grounding \method{} in computation complexity proxies.}
Each cell reports the correlation between the \method{} ranking and a ranking induced by a classical measure (output length, space, or time), shown as Spearman’s $\rho$ \;/\; Kendall’s $\tau_b$.
}
\resizebox{0.5\columnwidth}{!}{%
\begin{tabular}{l c c c}
\toprule
Model & Output & Space & Time \\
\midrule
Nemotron-49B  & 0.598\;/\;0.388 & 0.271\;/\;0.201 & 0.141\;/\;0.067 \\
Nemotron-253B & 0.537\;/\;0.356 & 0.166\;/\;0.101 & 0.137\;/\;0.101 \\
Gpt-oss-20B   & 0.665\;/\;0.482 & 0.518\;/\;0.422 & 0.471\;/\;0.382 \\
Gpt-oss-120B  & 0.606\;/\;0.440 & 0.356\;/\;0.253 & 0.362\;/\;0.287 \\
\midrule
Average       & 0.602\;/\;0.417 & 0.328\;/\;0.244 & 0.278\;/\;0.209 \\
\bottomrule
\end{tabular}%
}
\vspace{4pt}

\label{tab:grounding_traditional_complexity}
\end{table}

%% file: sections/appendix/SFT_detail.tex
\section{Supervised fine-tuning and inference details}
\label{appendix:SFT_detail}

\subsection{Supervised fine-tuning (SFT)}
\paragraph{Teacher supervision.}
We fine-tune \textsc{Qwen3-1.7B}  on \textsc{NVIDIA RTX A6000} with $48$ GB VRAM using supervised traces produced by \textsc{Nemotron-253B}. For each training instance, we use the teacher’s first-phase reasoning trace as the supervision target.

\paragraph{Chat formatting.}
Each example is serialized into a single text sequence using the model’s chat template:
\[
[\text{system}]~\Vert~[\text{user: prompt}]~\Vert~[\text{assistant: completion}],
\]
with a fixed system prompt:
\texttt{``Please solve the following algorithmic problem without using programming languages.''}

\paragraph{Parameter-efficient tuning (LoRA).}
We fine-tune using LoRA adapters with
$r=16$, $\alpha=32$, dropout $=0.05$, \texttt{bias=none}, and \texttt{target\_modules=all-linear}.
The base model is loaded with \texttt{torch\_dtype=bfloat16} and \texttt{device\_map=auto}, and we enable TF32 matmul for faster training.

\paragraph{Optimization hyperparameters.}
We train for 2 epochs with learning rate $2\times 10^{-4}$, per-device batch size $1$, and gradient accumulation $16$ (effective batch size $16$).
We use a cosine learning-rate schedule with warmup ratio $0.03$.
The maximum sequence length is set to 4096 tokens.

\subsection{Curriculum sampling used in SFT}
\paragraph{Temperature-annealed weighted sampling.}
In curriculum mode, we sample training examples with replacement from a temperature-controlled distribution over examples:
\[
P_t(i)\ \propto\ (1-\epsilon)\exp\!\left(-\mathcal{C}_i/\tau(t)\right) + \epsilon\cdot\frac{1}{N},
\]
where $\mathcal{C}_i$ is the example complexity (from \texttt{difficulty.json}), $N$ is dataset size, and $\epsilon$ is an exploration coefficient.
We linearly anneal the temperature from $\tau_0=0.15$ to $\tau_1=0.6$ across training steps, with $\epsilon=0.05$.

\subsection{Inference and evaluation (vLLM)}
\label{appendix:inference_detail}

\paragraph{Decoding configuration.}
We use greedy decoding by default (temperature $=0.0$, top-$p=1.0$) with repetition penalty $1.1$.
We set \texttt{max\_new\_tokens}=8192 and \texttt{max\_model\_len}=8192 to bound generation length and total context length.

\subsection{Key hyperparameters (summary)}
\begin{table}[H]
\centering
\small
\setlength{\tabcolsep}{6pt}
\caption{SFT and inference configurations used in our experiments.}
\begin{tabular}{l l}
\toprule
\textbf{Component} & \textbf{Setting} \\
\midrule
SFT base model & \textsc{Qwen3-1.7B} \\
SFT supervision & \textsc{Nemotron-253B} trace \\
SFT epochs / LR & 2 / $2\times 10^{-4}$ \\
SFT batch / accum & 1 / 16 \\
SFT max seq len & 4096 \\
SFT LR schedule & cosine, warmup ratio 0.03 \\
LoRA & $r=16$, $\alpha=32$, dropout 0.05, \texttt{all-linear} \\
Curriculum & $\tau_0=0.15$, $\tau_1=0.6$, $\epsilon=0.05$ \\
\midrule
Inference engine & vLLM (batched decoding) \\
Prompt mode & chat template, \texttt{enable\_thinking=True} \\
Decoding & greedy ($T=0$), top-$p=1.0$, rep.\ penalty 1.1 \\
Max tokens / context & \texttt{max\_new\_tokens}=8192, \texttt{max\_model\_len}=8192 \\
vLLM runtime & bf16, \texttt{gpu\_mem\_util}=0.90, batch size 32 \\
Eval filter & len $\in[10,100]$, idx $\in[16,19]$, len mod 10 = 0 \\
\bottomrule
\end{tabular}

\label{tab:sft_infer_hparams}
\end{table}

\subsection{Effectiveness of SFT}
\label{appendix:effectiveness_of_SFT}
\begin{figure}[H]
    \centering
    \includegraphics[width=\linewidth]{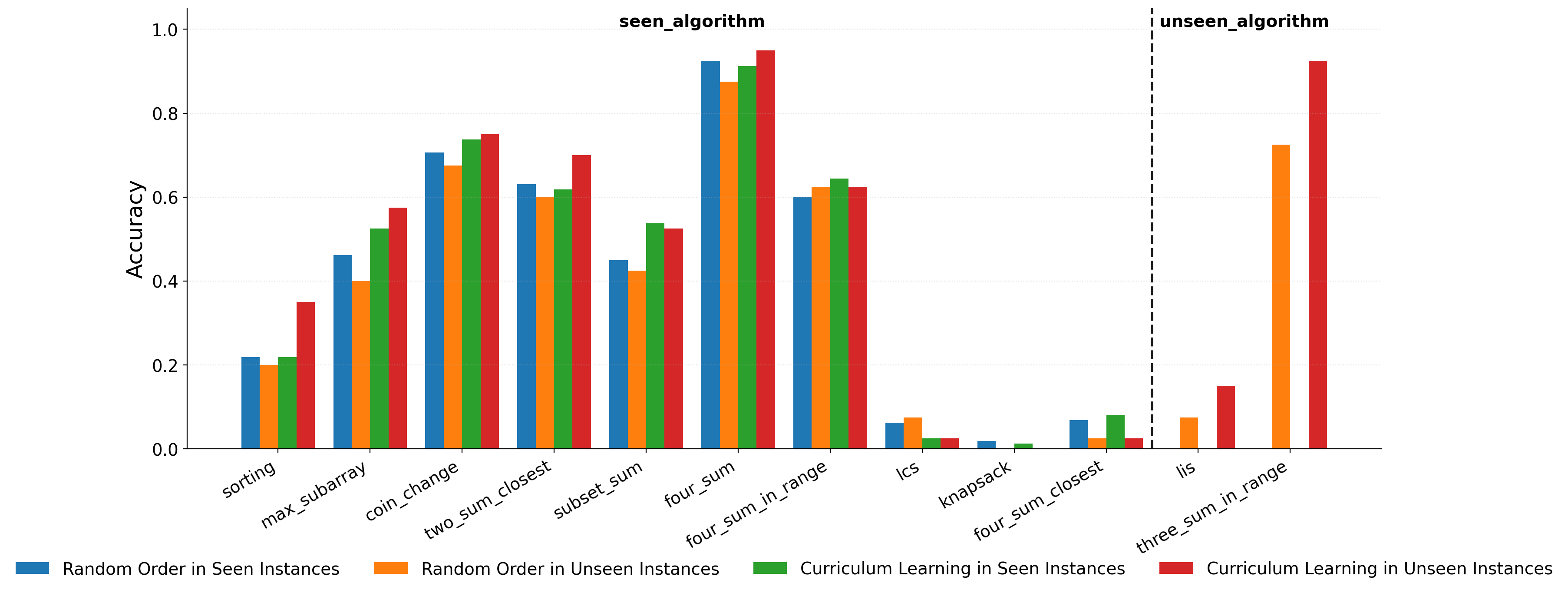}
    \caption{\textbf{SFT learns and curriculum improves generalization.}
    Accuracy on \emph{seen} (training-split) vs.\ \emph{unseen} (held-out) instances for each algorithm.
    Bars compare random-order SFT and curriculum SFT (complexity-based weighted sampling) evaluated on both splits.
    The dashed line separates algorithms included in SFT training (seen algorithms) from algorithms excluded from training (unseen algorithms).
    SFT yields higher accuracy on seen instances than unseen instances for seen algorithms, confirming effective adaptation, while curriculum SFT improves held-out performance and transfers better to unseen algorithms.}
    \label{fig:seen_vs_unseen}
\end{figure}

Figure~\ref{fig:seen_vs_unseen} validates that our SFT setup learns non-trivial signal from the teacher traces.
Across almost all \emph{seen} algorithms (left of the dashed line), the SFT model achieves substantially higher accuracy on \emph{seen instances} (training split) than on \emph{unseen instances} (held-out split), indicating that fine-tuning meaningfully adapts the base model to this problem family rather than producing random or degenerate outputs.
Moreover, curriculum-based sampling consistently improves performance over random-order SFT on the held-out split for most algorithms, suggesting better generalization instead of merely overfitting to the training instances.
Finally, on \emph{unseen algorithms} (right of the dashed line), curriculum SFT yields noticeably stronger transfer than random-order SFT, supporting our claim that the curriculum improves robustness to distribution shift.

\newpage